\pdfoutput=1
\documentclass[sigconf]{acmart}
\usepackage{multirow}

\AtBeginDocument{%
  }

\copyrightyear{2025}
\acmYear{2025}
\setcopyright{acmlicensed}
\acmConference[MM '25]{Proceedings of the 33rd ACM International Conference on Multimedia}{October 27-31, 2025}{Dublin, Ireland}

\acmBooktitle{Proceedings of the 33rd ACM International Conference on
Multimedia (MM '25), October 27--31, 2025, Dublin, Ireland}
\acmDOI{10.1145/3746027.3755774}
\acmISBN{979-8-4007-2035-2/2025/10}

\usepackage{multirow}
\usepackage{booktabs} 
\usepackage{amsmath} 

\begin{document}

\title{SAT: Supervisor Regularization and Animation Augmentation \\
for Two-process Monocular Texture 3D Human Reconstruction}


\author{Gangjian Zhang}
\affiliation{%
  \institution{HKUST(GZ)}
  \city{Guangzhou}
  \country{China}
}
\email{gzhang292@connect.hkust-gz.edu.cn}

\author{Jian Shu}
\affiliation{%
  \institution{HKUST(GZ)}
  \city{Guangzhou}
  \country{China}
}
\email{jshu704@connect.hkust-gz.edu.cn}

\author{Nanjie Yao}
\affiliation{%
  \institution{HKUST(GZ)}
  \city{Guangzhou}
  \country{China}
}
\email{nanjiey@uci.edu}



\author{Hao Wang}
\authornote{Corresponding author. \\
Project page: \url{https://sat-mm25.github.io/}.}
\affiliation{%
  \institution{HKUST(GZ)}
  \city{Guangzhou}
  \country{China}
}
\email{haowang@hkust-gz.edu.cn}


\renewcommand{\shortauthors}{Gangjian Zhang, Jian Shu, Nanjie Yao, \& Hao Wang}

\begin{abstract}
Monocular texture 3D human reconstruction aims to create a complete 3D digital avatar from just a single front-view human RGB image. However, the geometric ambiguity inherent in a single 2D image and the scarcity of 3D human training data are the main obstacles limiting progress in this field. To address these issues, current methods employ prior geometric estimation networks to derive various human geometric forms, such as the SMPL model and normal maps. However, they struggle to integrate these modalities effectively, leading to view inconsistencies, such as facial distortions. To this end, we propose a two-process 3D human reconstruction framework, SAT, which seamlessly learns various prior geometries in a unified manner and reconstructs high-quality textured 3D avatars as the final output. To further facilitate geometry learning, we introduce a Supervisor Feature Regularization module. By employing a multi-view network with the same structure to provide intermediate features as training supervision, these varied geometric priors can be better fused. To tackle data scarcity and further improve reconstruction quality, we also propose an Online Animation Augmentation module. By building a one-feed-forward animation network, we augment a massive number of samples from the original 3D human data online for model training. Extensive experiments on two benchmarks show the superiority of our approach compared to state-of-the-art methods. 
\end{abstract}



\begin{teaserfigure}
  \includegraphics[width=1\linewidth]{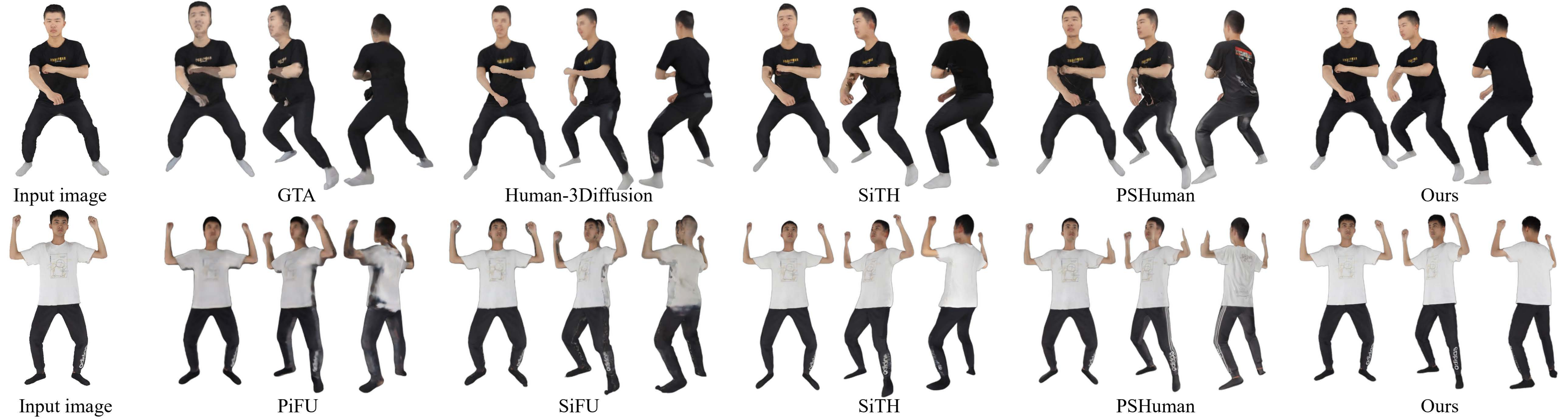}
    \vspace{-0.7cm}
  \caption{Qualitative Comparison with SOTA Methods. For the 3D result of each method, we render them from three views. Comparison shows our method can achieve better human reconstruction quality, and produces less blurring and deformities.}
  \label{fig:teaser}
\end{teaserfigure}





\maketitle

\section{Introduction}
\label{sec:intro}
Texture 3D human reconstruction is a challenging task that involves reconstructing the entire 3D avatar from a monocular frontal RGB human image. This task has gained attention due to its applications in VR/AR. However, since the input is a single 2D image, there is an inherent ambiguity in geometry perception. Unlike stereo vision, where depth information can be inferred from the disparity between two images, monocular vision lacks this direct depth cue.

Previous methods have typically relied on various geometric estimation networks to obtain different modality forms of human geometric priors, such as rough 3D body meshes~\cite{ho2024sith, VS_CVPR2024, hilo} (using the SMPL model~\cite{loper2023smpl, pavlakos2019expressive_smplx, romero2022embodied_smplh}), human normal maps~\cite{saito2019pifu, xiu2022icon, xiu2023econ, Zhang_2024_sifu}, or depth maps~\cite{tang2025hap}. However, these methods struggle to effectively integrate information from these diverse sources, leading to notable inconsistencies in the resulting views, such as distortions in the side faces. Additionally, although the complexity of human poses surpasses that of generic object reconstruction~\cite{szymanowicz2023splatter, tang2024lgm, xu2024grm, xu2024agg, zhang2024gslrm}, there is a more severe scarcity of publicly available 3D scan data of humans~\cite{tao2021function4d_thuman} compared to the object domain~\cite{chang2015shapenetinformationrich3dmodel, deitke2023objaverse, objaverseXL}. This limitation further hampers the performance of these methods in this domain.

We propose a two-process reconstruction framework, SAT, to achieve more effective human reconstruction. This framework contains four components: United Geometry Learning (UGL), Supervisor Feature Regularization (SFR), Cascading Gaussian Texturing (CGT), and Online Animation Augmentation (OAA).



In the initial phase, we introduce UGL, which integrates multi-modal geometric information from various pre-existing models into a unified geometric learning network to enhance 3D human geometry Gaussian reconstruction. To support and fully integrate these priors, we propose an SFR module. This module involves training a supervisor model that mirrors the structure of the monocular reconstruction network. It uses multi-view ground truth (GT) geometry to generate pseudo-GT intermediate features, guiding the learning of corresponding features in the monocular reconstruction network and improving reconstruction quality.

In the second phase, we conduct CGT, which directly uses the 3D geometry Gaussian inferred from the first phase, along with the input image, to reconstruct the 3D texture Gaussian. This ensures that the second phase aligns with the output distribution from the first phase, minimizing cascading errors. To enhance reconstruction quality and address the shortage of 3D human data, we incorporate an OAA module for online sample augmentation. We train an animation network based on an existing animation dataset~\cite{shen2023xavatar}, which efficiently performs pose transformation of existing 3D human data in a single online forward propagation, generating numerous augmented samples for model training.

We validated the proposed method's ability to achieve SOTA performance on two publicly available test datasets, CustomHuman~\cite{ho2023customhuman} and THuman3.0~\cite{thuman3.0}. Overall, our contribution can be summarized:
\begin{itemize}

    \item A two-process monocular texture 3D human reconstruction framework is proposed, which introduces two sequential processes, United Geometry Learning and Cascading Gaussian Texturing, to achieve effective human reconstruction.

    \item A Supervisor Feature Regularization module is proposed, which considers training a supervisor model to regularize human geometry learning towards more accurate directions.
    \item An online animation augmentation module is proposed, which involves training an animation model to augment 3D human data online, providing more samples for model training.

\end{itemize}

\section{Related Work}
\label{sec:rw}

\subsection{Monocular 3D Human Reconstruction}

Monocular 3D human reconstruction begins with PIFu~\cite{saito2019pifu}, which introduces pixel-aligned implicit functions for shape and texture. ICON~\cite{xiu2022icon} enhances this by employing skinned body models~\cite{loper2023smpl} as body priors, while ECON~\cite{xiu2023econ} integrates implicit representations with explicit body regularization.  GTA~\cite{zhang2024global_gta} employs a 3D-decoupling transformer for detailed reconstruction. VS~\cite{VS_CVPR2024} introduces a "stretch-refine" strategy to handle large deformations in loose clothing. HiLo~\cite{hilo} improves geometry detail by extracting high and low-frequency information from a parametric model, enhancing noise robustness. SiTH~\cite{ho2024sith} addresses occlusion using a 2D prior from the SD model. SiFU~\cite{Zhang_2024_sifu} uses cross-attention for geometry optimization and 2D diffusion for texture. Human-3Diffusion~\cite{3diffusion} proposes joint training for enhanced human reconstruction. PSHuman~\cite{li2024pshuman} introduces a blending diffusion method for improved facial and body reconstruction quality. MultiGo~\cite{zhang2024multigo} aims to enhance human geometry across multiple human levels.
 

\subsection{3D Human Gaussian}

Recent advancements in 3D Gaussian splatting~\cite{3DGaussian} have enhanced 3D human modeling, while traditional methods like SDF~\cite{sdf} and NeRF~\cite{mildenhall2021nerf} often struggle with efficiency and rendering quality. Innovative techniques such as HuGS~\cite{HuGS}, D3GA~\cite{D3GA}, and 3DGS-Avatar~\cite{3DGS-Avatar} utilize spatial and temporal data for modeling from various video inputs. Gauhuman~\cite{hu2024gauhuman} and HUGS~\cite{HuGS} optimize human Gaussians from monocular videos, HiFi4G~\cite{Hifi4g} ensures spatial-temporal consistency with a dual-graph mechanism, and ASH~\cite{Ash} uses mesh UV parameterization for real-time rendering. Animatable Gaussians and GPS-Gaussian~\cite{GPS-Gaussian} enhance dynamics and quality.

\subsection{3D Human Avatar Animation}  

Current animation methods are often based on LBS, focusing on solving the artifact problem caused by directly applying LBS transformation. Various methods~\cite{ho2023llevh, shen2023xavatar, paudel2024ihuman, Qian_2024_3dgsavatar, moon2024expressive, wen2024gomavatarefficientanimatablehuman, yuan2024gavatar, karthikeyan2024avatarone} typically optimize an animatable template from human data (single or multi-view) and integrate it with body models~\cite{loper2023smpl,scape2005}. Animation is achieved by transforming the canonical space to the target space using LBS. While these optimization-based approaches produce high-quality results, they are time-intensive. Recent advancements~\cite{wen2025life} have improved efficiency, yet the process still takes at least 1 second.

\begin{figure*}
    \centering
    \includegraphics[width=0.99\linewidth]{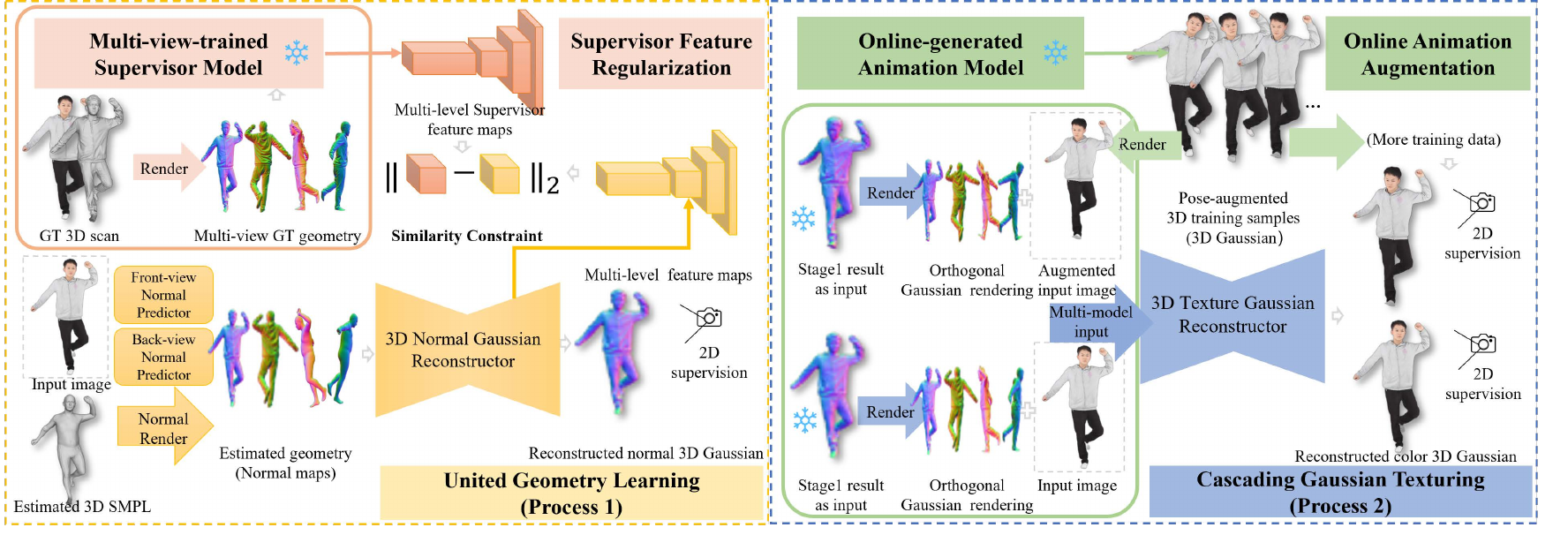}
    \vspace{-0.45cm}

    \caption{\textbf{Method Overview.} We introduce SAT for monocular texture 3D human reconstruction, comprising two processes: United Geometry Learning (UGL) and Cascading Gaussian Texturing (CGT). Central to them are two modules: Supervisor Feature Regularization (SFR) and Online Animation Augmentation (OAA). In the UGL process, we extract multi-modal geometric features from the input image using various prior models and integrate them into a unified network to reconstruct 3D human normal Gaussians. The SFR enhances this by training a multi-view supervisor model that creates feature maps, serving as a regularizing force for monocular geometry learning. The CGT process focuses on aligning with the output distribution from UGL while minimizing cascading errors. It utilizes the output from UGL and the input image to reconstruct the 3D texture Gaussian. To improve reconstruction quality and address the scarcity of 3D human data, we introduce OAA, which trains an animation model to dynamically generate diverse pose samples, augmenting the dataset for better model training.}
    
    \label{fig: pipeline}
    \vspace{-0.1cm}
\end{figure*}

\section{Methodology}
\label{sec:method}

\subsection{Preliminaries}\label{prel}
\textbf{Gaussian Splatting.} Gaussian Splatting \cite{3DGaussian} is a popular 3D representation method using a set of 3D Gaussians, $\Theta$, to model 3D data. Each Gaussian is defined by a parameter set: $\theta_i = {\mathbf{x}_i, \mathbf{s}_i, \mathbf{q}_i, \alpha_i, \mathbf{c}_i} \in \mathbb{R}^{14}$. Here, $\mathbf{x} \in \mathbb{R}^3$ represents the geometry center, $\mathbf{s} \in \mathbb{R}^3$ the scaling factor, $\mathbf{q} \in \mathbb{R}^4$ the rotation quaternion, $\alpha \in \mathbb{R}$ the opacity value, and $\mathbf{c} \in \mathbb{R}^3$ the color feature.




\noindent \textbf{SMPL Series Model.} The SMPL (Skinned Multi-Person Linear) model, introduced in work~\cite{loper2023smpl}, is a parametric framework designed to represent the 3D human body. It employs body parameters denoted as $\beta \in \mathbb{R}^{k}$ to define a specific human body mesh $\mathcal{M}(\beta) \in \mathbb{R}^{3\times6890}$. Each parameter within $\beta$ influences the position/orientation of body parts. Several extensions~\cite{pavlakos2019expressive_smplx, romero2022embodied_smplh, zhang2023pymaf, pixie}, incorporate additional parameters to capture facial expressions, finger movements, and more. In this paper, we primarily utilize SMPL-X~\cite{pavlakos2019expressive_smplx} as our human body model.

\noindent \textbf{Linear Blend Spinning (LBS).} LBS is a technique for transforming a human 3D scan from a original pose to a new target pose based on the original and target SMPL models. It calculates the position of the deformed vertex by taking a weighted average of the transformations of each influencing bone. In this paper, LBS serves as the baseline method for transforming human scans to generate augmented 3D human training samples online.

\subsection{Overview}


Figure~\ref{fig: pipeline} provides a comprehensive overview of our proposed approach, highlighting the key components and their interactions. Section~\ref{sec:UGL} introduces the United Geometric Learning (UGL) method, while Section~\ref{sec:SFR} describes the Supervisor Feature Regularization (SFR) module. In Section~\ref{sec:CGT}, we present the Cascading Gaussian Texturing (CGT) process, and finally, Section~\ref{sec:OAA} introduces the Online Animation Augmentation (OAA) module.

\subsection{United Geometry Learning} \label{sec:UGL}
In monocular texture 3D human reconstruction, the challenge lies in reconstructing accurate posture and depth from just a front-view image. Traditional methods have utilized various geometric estimation networks to derive multiple human geometric priors, like 3D body meshes (e.g., SMPL model), normal maps, and depth maps. However, these approaches often fail to integrate this diverse information effectively, resulting in inconsistencies, such as distorted side profiles. To tackle these issues, we propose a United Geometry Learning (UGL) method that enables the interactive fusion of geometric priors from multiple sources, addressing these issues.

As shown in Figure~\ref{fig: pipeline}, we consider three different sources of human geometry prior, one front-view normal predictor, and one back-view normal predictor, as well as a 3D body mesh estimator. We input a human front-view RGB image $\mathcal{I} \in \mathbb{R}^{3 \times H \times W}$ into these three models to obtain the front-view and back-view normals $\widehat{\mathcal{N}}^{\mathrm{f}}$, $\widehat{\mathcal{N}}^{\mathrm{b}} \in \mathbb{R}^{3 \times H \times W}$, and the SMPL 3D body mesh $\mathcal{S}$ which is subsequently rendered from the left and right sides to also transform into the normal modality ($\widehat{\mathcal{N}}^{\mathrm{l}}$, and $\widehat{\mathcal{N}}^{\mathrm{r}}$ $\in \mathbb{R}^{3 \times H \times W}$). Since these normal maps come from different sources, they are not view-inconsistent. 

To solve this, we concatenate and input them ($[\widehat{\mathcal{N}}^{\mathrm{f}},\widehat{\mathcal{N}}^{\mathrm{b}}, \widehat{\mathcal{N}}^{\mathrm{l}}, \widehat{\mathcal{N}}^{\mathrm{r}}] \in \mathbb{R}^{4 \times 3 \times H \times W}$) into a 3D Gaussian reconstructed UNet~\cite{tang2024lgm}, which containing cross-view self-attention layers. The predicted 3D Gaussian is supervised by view-consistent normal maps rendered from human GT scans. This attention mechanism, combined with view-consistent supervision, facilitates effective interaction and smoothing at the edges of normal maps of different views. We follow previous work~\cite{tang2024lgm} by comparing the differentiable rendering~\cite{3DGaussian} from the predicted 3D human normal Gaussian with the Nvdiffrast~\cite{nvdiffrast} rendering from the GT human scans:
\begin{equation} \label{equa1}
\begin{split} 
    & \mathcal{L_{\mathrm{1}}}=\sum_{i=1}^{V} \left( \mathcal{L}_{\mathrm{mse}}\left(N_{\mathrm{i}}, {{N_{\mathrm{i}}}^{\mathrm{GT}}; M_{\mathrm{i}},{M_{\mathrm{i}}}^{\mathrm{GT}}}\right) + \mathcal{L}_{\mathrm{lpips}}\left(N_{\mathrm{i}}, {N_{\mathrm{i}}}^{\mathrm{GT}}\right) \right),
\end{split}
\end{equation}
where $i$ is the $i^{th}$ camera rendering view, $V$ is the number of rendering views. $N_{\mathrm{i}}$ and $M_{\mathrm{i}} \in \mathbb{R}^{3 \times H \times W}$ are the image and mask rendered from the predicted Gaussian. ${N_{\mathrm{i}}}^{\mathrm{GT}}$ and ${M_{\mathrm{i}}}^{\mathrm{GT}} \in \mathbb{R}^{3 \times H \times W}$ are the normal map and mask rendered from the GT scans. We use the MSE loss to compute the pixel-level difference and the LPIPS~\cite{lpips} loss to compute the semantic-level difference.







\begin{figure}
    \centering

    \includegraphics[width=0.99\linewidth]{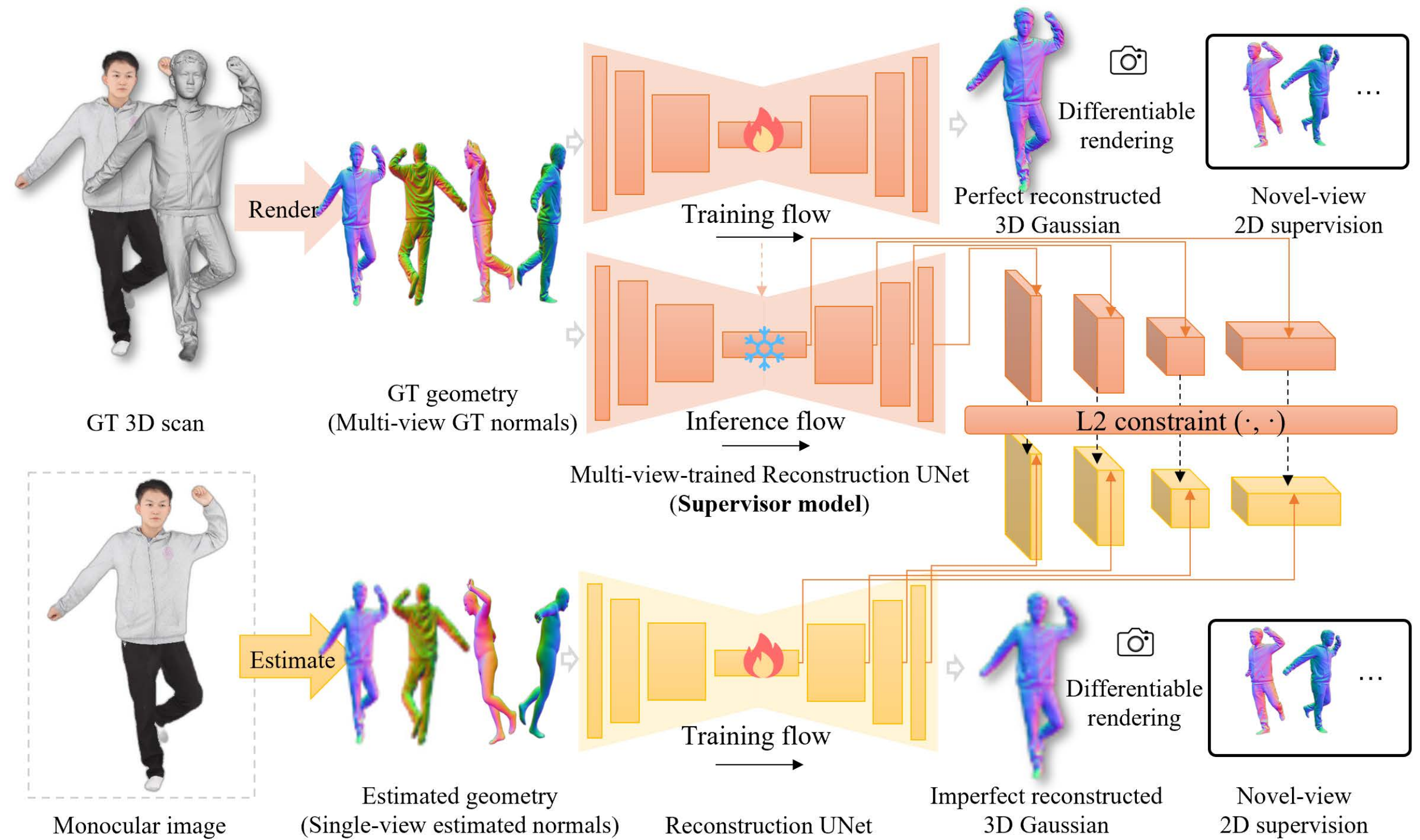}
    \vspace{-0.3cm}

    \caption{\textbf{Supervisor Feature Regularization.} To better fuse the inaccurate geometric priors from various sources, we propose building a supervisor model (orange UNet) that shares the same structure and is trained with multi-view GT human normal maps. This model extracts optimal intermediate features for 3D reconstruction, offering enhanced training guidance for the monocular geometry learning process.}
    \label{fig: SFR}
    \vspace{-0.2cm}

\end{figure}


\subsection{Supervisor Feature Regularization} \label{sec:SFR}

In UGL, we integrate geometric information from various prior models to create a foundational 3D human geometry Gaussian. However, the accuracy of this geometry is limited by the performance of these models, which may lead to flawed details. To enhance the integration of geometric information, we propose a Supervisor Feature Regularization (SFR) module to improve the overall geometry learning process. As shown in Figure~\ref{fig: SFR}, SFR module employs a supervisor model that mirrors the architecture of the reconstruction UNet from the UGL (in Section~\ref{sec:UGL}). The main difference is that the supervisor network is trained on complete GT geometric information, specifically four orthogonal human normal maps from GT scans. This allows the supervisor to generate an accurate 3D human Gaussian, with features from the UNet's hidden layers closely aligned with the ground truth. These supervisor features are then used to guide and regularize our monocular geometry learning process.

Concretely, to train the supervisor model, we firstly render multi-view normal maps (front, back, left, right) directly from the GT human scan to obtain $[\mathcal{N}^{\mathrm{f}},\mathcal{N}^{\mathrm{b}},\mathcal{N}^{\mathrm{l}},\mathcal{N}^{\mathrm{r}}] \in \mathbb{R}^{4 \times 3 \times H \times W}$. We then feed them into the reconstruction UNet to predict 3D normal Gaussian, and supervise it with the normal maps rendered by the same human scan. The supervisor training constraint is similar to Equation~\ref{equa1}, and we use the MSE and LPIPS losses to compute the rendering difference from the reconstructed Gaussian and GT scan: 
\begin{equation} \label{equaS}
\begin{split} 
    & \mathcal{L_{\mathrm{S}}}=\sum_{j=1}^{V} \left( \mathcal{L}_{\mathrm{mse}}\left({N^{*}_{\mathrm{j}}}, {{N_{\mathrm{j}}}^{\mathrm{GT}}; {M^{*}_{\mathrm{j}}},{M_{\mathrm{j}}}^{\mathrm{GT}}}\right) + \mathcal{L}_{\mathrm{lpips}}\left({{N^{*}_{\mathrm{i}}}}, {N_{\mathrm{i}}}^{\mathrm{GT}}\right) \right),
\end{split}
\end{equation}
where $j$ is the $j^{th}$ camera rendering view, $V$ is the number of rendering views. ${N^{*}_{\mathrm{j}}}$ and ${M^{*}_{\mathrm{j}}} \in \mathbb{R}^{3 \times H \times W}$ are the image and mask rendered from the 3D human normal Gaussian from multi-view input. ${N_{\mathrm{j}}}^{\mathrm{GT}}$ and ${M_{\mathrm{j}}}^{\mathrm{GT}} \in \mathbb{R}^{3 \times H \times W}$ are the normal map and mask from the GT scan. Since the multi-view GT geometry input is used for training, the supervisor model can reconstruct a nearly perfect 3D human.

To use this supervisor model to assist us in monocular reconstruction, we first freeze the parameters of the trained supervisor model. Besides the procedure in section~\ref{sec:UGL}, we will additionally render human normal maps ($\mathcal{N}^{\mathrm{f}}$, $\mathcal{N}^{\mathrm{b}}$, $\mathcal{N}^{\mathrm{l}}$, $\mathcal{N}^{\mathrm{r}} \in \mathbb{R}^{3 \times H \times W}$) from the human GT scan for the current training sample. They are fed into the supervisor model to infer and obtain the feature maps of the middle and up blocks $\{\mathcal{F}_{k}^{\mathrm{S}}|k=\{mid,up1,up2,up3,up4,up5\}\}$ (The UNet contain one middle and five up blocks). Similarly, we output the feature maps from the same blocks from the monocular reconstruction UNet to obtain $\{\mathcal{F}_{k}|k \in \{mid,up1,up2,up3,up4,up5\}\}$. We then introduce our supervisor feature regularization constraint:
\begin{equation} \label{equaSFR}
    \mathcal{L}_{SFR} = \sum_{k} \Vert {\mathcal{F}_{\mathrm{k}}}^{\mathrm{S}} - \mathcal{F}_{\mathrm{k}} \Vert_{2}.
\end{equation}
Since the supervisor's feature maps better encapsulate the features required for perfect reconstruction, here, we bring the intermediate features of monocular reconstruction UNet closer to those of multi-view reconstruction UNet, to force the former to learn these features in these intermediate network layers. Once these features are learned, they can naturally reconstruct a better 3D human.

\subsection{Cascading Gaussian Texturing  
} \label{sec:CGT}

After reconstructing the geometric 3D human Gaussian, we focus on obtaining the final color 3D human avatar in the subsequent process. To enhance the quality of this reconstruction, we do not treat these two processes as separate and independent. As shown in Figure~\ref{fig: pipeline}, we utilize the inference output from the first process model as the direct input for the second process. This allows the training of the second process to align closely with the output distribution of the first process, mirroring the settings of our method during inference. We refer to this method as Cascading Gaussian Texturing (CGT). By incorporating inputs from both modalities—the input image and the 3D normal Gaussian output from the first process. CGT can more effectively reconstruct the color 3D human avatar.

Concretely, we infer the 3D human normal Gaussian from process one. Subsequently, we perform Gaussian rendering on it from four orthogonal views (front, back, left, right) to obtain four images, $\mathcal{I}_{\mathcal{G}}^{f}$, $\mathcal{I}_{\mathcal{G}}^{b}$, $\mathcal{I}_{\mathcal{G}}^{l}$, $\mathcal{I}_{\mathcal{G}}^{r}$ $\in \mathbb{R}^{3 \times H \times W}$. They are concatenated with the input image ($[\mathcal{I}_{\mathcal{G}}^{f}, \mathcal{I}_{\mathcal{G}}^{b},\mathcal{I}_{\mathcal{G}}^{l},\mathcal{I}_{\mathcal{G}}^{r},\mathcal{I}] \in \mathbb{R}^{5 \times3 \times H \times W}$) and fed to a texture Gaussian reconstruction UNet with the same structure in Geometry reconstruction (Section~\ref{sec:UGL}). Five images from two different modalities will be interacted and fused through the attention layers in UNet, where the Gaussian modality conveys the human geometric information to the RGB modality, while the latter conveys non-frontal-view texture information to the former. Predicted 3D color Gaussian supervised by view-consistent color images rendered from the human GT scan by MSE and LPIPS losses:
\begin{equation} 
\label{equa2}
\begin{split} 
    & \mathcal{L_{\mathrm{2}}}=\sum_{m=1}^{V} \left( \mathcal{L}_{\mathrm{mse}}\left(I_{\mathrm{m}}, {I_{\mathrm{m}}^{\mathrm{GT}}; M_{\mathrm{m}},M_{\mathrm{m}}^{\mathrm{GT}}}\right) + \mathcal{L}_{\mathrm{lpips}}\left(I_{\mathrm{m}}, I_{\mathrm{m}}^{\mathrm{GT}}\right) \right),
\end{split}
\end{equation}
where $m$ is the $m^{th}$ camera rendering view, $V$ is the number of rendering views. $I_{\mathrm{m}}$ and $M_{\mathrm{m}}\in \mathbb{R}^{3 \times H \times W}$ are the image and mask rendered from the predicted Gaussian. ${I_{\mathrm{m}}}^{\mathrm{GT}}$ and ${M_{\mathrm{m}}}^{\mathrm{GT}}\in \mathbb{R}^{3 \times H \times W}$ are the RGB image and mask rendered from the GT human scan.

\textbf{It is worth noting} that the cascading training we propose is compared to separate training, referring to using orthogonal normal maps rendered directly from the GT human scan as input during training, instead of the predicted 3D Gaussian from the first process.





\subsection{Online Animation Augmentation 
} \label{sec:OAA}

\begin{figure}
    \centering

    \includegraphics[width=1\linewidth]{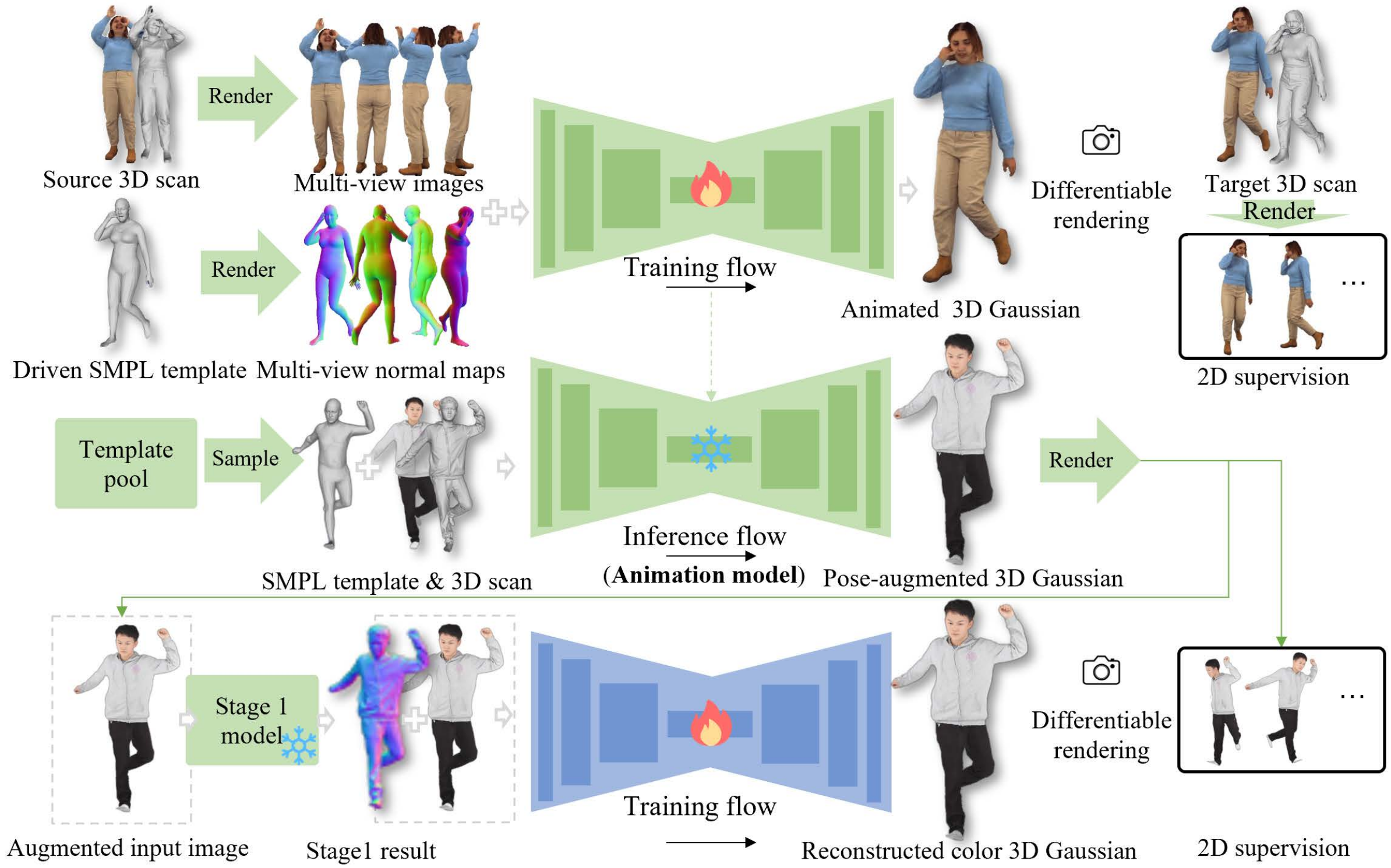}
        \vspace{-0.7cm}
    \caption{Online Animation Augmentation. To provide more diverse samples for training, we propose training a 3D human animation network (green UNet), which can infer and transform original 3D human data into specified poses based on dynamic pose templates. This will enable the generation of a large number of new augmented 3D samples online.}
    \label{fig:OAA}
    \vspace{-0.3cm}
\end{figure}


Through the above operations, we derive 3D texture human. However, the limited availability of 3D human scan datasets restricts reconstruction performance. To address this, we introduce Online Animation Augmentation (OAA) module. As shown in Figure~\ref{fig:OAA}, our method trains a human animation model to generate new 3D human samples in specific poses using existing scan data and a driving pose template. This enables online data augmentation for training. Unlike other 3D human animation methods that require significant time for results, our approach uses a feedforward network~\cite{tang2024lgm} for real-time data generation, greatly increasing the volume of augmented training data. Additionally, unlike the LBS method discussed in Section~\ref{prel}, which also meets real-time needs, our approach reduces distortion and yields higher-quality training samples.


Before training our animation model, we prepare the training data in advance, which is a triplet ($\mathcal{S}^{o}$, $\mathcal{M}^{t}$, $\mathcal{S}^{t}$), where $\mathcal{S}^{o}$ the original human scan, $\mathcal{M}^{t}$ is the driven SMPL template denoting the target pose wanted to be transformed, and $\mathcal{S}^{t}$ is the target human scan. In Figure~\ref{fig:OAA}, we rendered four orthogonal RGB images from $\mathcal{S}^{o}$ and four orthogonal normal maps from $\mathcal{M}^{t}$ under the same camera views. We concatenate and feed them ($\in \mathbb{R}^{8 \times 3 \times H \times W}$) into a 3D Gaussian reconstruction network, with the same structure in Section~\ref{sec:CGT}, to fuse two different modalities and predict the target-pose 3D human color Gaussian, which is supervised by color images rendered from the target scan using the same manner in Equation~\ref{equa2}:
\begin{equation} 
\label{equaA}
\begin{split} 
    & \mathcal{L_{\mathrm{A}}}=\sum_{n=1}^{V} \left( \mathcal{L}_{\mathrm{mse}}\left(I^{*}_{\mathrm{n}}, {I_{\mathrm{n}}^{\mathrm{GT}}; M^{*}_{\mathrm{n}},M_{\mathrm{n}}^{\mathrm{GT}}}\right) + \mathcal{L}_{\mathrm{lpips}}\left(I^{*}_{\mathrm{n}}, I_{\mathrm{n}}^{\mathrm{GT}}\right) \right),
\end{split}
\end{equation}
where $n$ the $n^{th}$ camera rendering view, $V$ is the number of rendering views. $I^{*}_{\mathrm{n}}$ and $M^{*}_{\mathrm{n}}$ are the image and mask rendered from the predicted target-pose 3D human color Gaussian. ${I_{\mathrm{n}}}^{\mathrm{GT}}$ and ${M_{\mathrm{n}}}^{\mathrm{GT}}$ are the RGB image and mask rendered from the target GT human scan.  After obtaining this trained animation model, we use it to augment the original 3D scan in the training dataset online. 

As shown in Figure~\ref{fig:OAA}, we first put all SMPL templates from the training dataset~\cite{tao2021function4d_thuman} (each human scan corresponds to a GT SMPL template) into a template pool. During training, we randomly select a pose template and feed it together with one existing 3D scan into our parameter-frozen animation model to infer a pose-augmented 3D Gaussian. Unlike the original training samples, which are presented as a 3D scan, we obtain a 3D Gaussian here. However, it can also be used as a training sample. We only need to render images from the 3D Gaussian as training input and supervision.

\begin{table*}[t!]
    \centering

        \caption{\textbf{3D Comparison with SOTA Methods.}An $\uparrow$ or $\downarrow$ arrow means a higher or lower value is preferable. All experiments were conducted using the estimated SMPL/SMPL-X model. An asterisk denotes that this particular method utilized the test set as training data for model training. Under the same condition, the \textbf{best} results are marked in bold.  \label{main_exp_3d}}

            \vspace{-0.1cm}

    \renewcommand{\arraystretch}{0.87}
    \scalebox{0.88}{
    \begin{tabular*}{\textwidth}{@{\extracolsep{\fill}}lc|ccc|ccc}
    \toprule
         \multirow{2}{*}{Methods}  & \multirow{2}{*}{Publication}     & \multicolumn{3}{c}{CustomHuman~\cite{ho2023customhuman}} & \multicolumn{3}{c}{THuman3.0~\cite{thuman3.0}}  \\
                                   &                               &  CD: P-to-S / S-to-P $(\mathrm{cm}) \downarrow$  & NC $\uparrow$ & f-score $\uparrow$ &
                                                                     CD: P-to-S / S-to-P $(\mathrm{cm}) \downarrow$  & NC $\uparrow$ & f-score $\uparrow$  \\
    \midrule 
    PIFu~\cite{saito2019pifu}        &  ICCV 2019  & $2.965/3.108$ & 0.765 & 25.708 & $2.176/2.452$  & 0.773 & 34.194 \\
    ICON~\cite{xiu2022icon}          & CVPR 2022   & $2.437/2.811$ & 0.783 & 29.044 & $2.471/2.780$   & 0.756 & 27.438 \\
    ECON~\cite{xiu2023econ}          & CVPR 2023   & $2.192/2.342$ & 0.806 & 33.287 & $2.200/2.269$   & 0.781 & 33.220 \\   
    GTA~\cite{zhang2024global_gta}   & NeurIPS 2023& $2.404/2.726$ & 0.790 & 29.907 & $2.416/2.652$   & 0.768 & 29.257 \\    
    VS~\cite{VS_CVPR2024}            & CVPR 2024   & $2.508/2.986$ & 0.779 & 26.790 & $2.523/2.943$   & 0.758 & 26.340  \\    
    HiLo~\cite{hilo}                 & CVPR 2024   & $2.280/2.739$ & 0.794 & 30.275 & $2.385/2.862$   & 0.765 & 28.119 \\
    SiFU~\cite{Zhang_2024_sifu}      & CVPR 2024   & $2.464/2.782$ & 0.790 & 28.574 & $2.480/2.822$   & 0.762 & 27.929 \\
    SiTH~\cite{ho2024sith}           & CVPR 2024   & $1.800/2.188$ & 0.816 &36.148 & $1.763/2.002$ &0.787  &36.230 \\

    PSHuman~\cite{li2024pshuman}                  & CVPR 2025   & $1.857/2.106$  & 0.832   & 37.899     & $1.803/2.059$   &   0.799   &   39.960      \\
    MultiGO~\cite{zhang2024multigo}               & CVPR 2025   & $1.620/1.782$  & 0.850   & 42.425     & $1.408/1.633$   &   0.834   &   46.091   \\

    Ours                    &     /      & $\textbf{1.546}/\textbf{1.657}$ & \textbf{0.857} & \textbf{44.436} & $\textbf{1.339}/\textbf{1.570}$ & \textbf{0.840} & \textbf{47.605} \\
    \midrule
        Human-3Diffusion*~\cite{3diffusion}  & NeurIPS 2024& $1.832/1.229$ & 0.818 & 43.245 & $1.799/1.122$   &   0.803   &   43.866   \\

    \bottomrule
    \end{tabular*}
    }
    \vspace{-0.2cm}

\end{table*}

\begin{table*}[t!]
    \centering

        \caption{\textbf{2D Comparison with SOTA Methods.} The 3D results of each method will be rendered 2D RGB images from the front and back views, represented by `F' and `B' in the table. Some methods, like ICON and ECON, only predict the front view texture.\label{main_exp_2d}}
        
            \vspace{-0.1cm}
    \renewcommand{\arraystretch}{0.87}
        \scalebox{0.88}{
    \begin{tabular*}{\textwidth}{@{\extracolsep{\fill}}l|ccc|ccc}
    \toprule
         \multirow{2}{*}{Methods}  & \multicolumn{3}{c}{CustomHuman} & \multicolumn{3}{c}{THuman3.0}  \\
                                   & LPIPS: F/B $\downarrow$ & SSIM: F/B $\uparrow$ & PSNR: F/B $\uparrow$ &
                                     LPIPS: F/B $\downarrow$ & SSIM: F/B $\uparrow$ & PSNR: F/B $\uparrow$   \\
    \midrule
    PIFu    & $0.0792/0.0966$ & $0.8965/0.8742$ & $18.141/16.721$ & $0.0706/0.0849$ & $0.9242/0.9007$ & $20.104/17.926$ \\  
    ICON    & $0.0710/-$ & $0.8976/-$ & $18.613/-$ & $0.0608/-$ & $0.9291/-$ & $21.127/-$\\     
    ECON    & $0.0781/-$ & $0.8868/-$ & $18.454/-$ & $0.0658/-$ & $0.9261/-$ & $20.961/-$ \\ 
    GTA     & $0.0730/0.0891$ & $0.9003/0.8923$ & $18.790/18.229$ & $0.0633/0.0770$ & $0.9298/0.9275$ & $21.113/20.497$ \\
    SiFU    & $0.0692/0.0879$ & $0.9023/0.8915$ & $18.715/18.111$ & $0.0597/0.0768$ & $0.9302/0.9243$ & $21.101/20.349$ \\
    SiTH    & $0.0679/0.0843$ & $0.9007/0.8870$ & $18.417/17.608$ & $0.0612/0.0766$ & $0.9232/0.9107$ & $20.326/19.355$ \\
    Human-3Diffusion & $0.0546/0.0639$ & $0.9405/0.9368$ & $21.909/20.736$ & $0.0532/0.0612$ & $0.9578/0.9501$ & $23.502/22.142$ \\
    PSHuman & $0.0612/0.0721$ & $0.9147/0.9131$ & $19.881/19.631$ & $0.0567/0.0689$ & $0.9402/0.9438$ & $21.852/21.246$ \\
MultiGO  & $0.0414/0.0643$ & $0.9603/0.9415$ & $22.347/20.849$ & $0.0457/0.0616$ & $0.9623/0.9512$ & $23.794/22.657$ \\

    Ours    & $\textbf{0.0358/0.0577}$ & $\textbf{0.9701/0.9499}$ & $\textbf{23.351/21.515}$ & $\textbf{0.0398/0.0588}$ & $\textbf{0.9738/0.9552}$ & $\textbf{25.181/23.119}$ \\  

    \bottomrule
    \end{tabular*}
    }
    \vspace{-0.2cm}

\end{table*}

\begin{table*}[t!]
    \centering
    \caption{\textbf{Ablation Study on Our Components.} In this table, from top to bottom, we continuously stack and introduce each component we propose. ``Our two-process framework” method refers to the two-process reconstruction framework proposed by us, which we consider as the baseline method here. ``+Supervisor regularization” means that we add the proposed Supervisor Feature Regularization (SFR) module on top of the baseline. In ``+Cascading training”, we further introduce the proposed Cascading Gaussian Texturing (CGT) operation in the second process training. In the last two rows of the table, we attempt to compare the impact of two online sample augmentation strategies, the proposed Online Animation Augmentation (OAA) method and the common LBS method, on model performance by introducing them separately.} \label{abl_exp_3d}
            \vspace{-0.1cm}

    \renewcommand{\arraystretch}{0.87}
    \scalebox{0.88}{
    \begin{tabular*}{\textwidth}{@{\extracolsep{\fill}}l|ccc|ccc}
    \toprule
    \multirow{2}{*}{Methods}  & \multicolumn{2}{c}{CustomHuman} & \multicolumn{3}{c}{THuman3.0}  \\
                                                    & CD: P-to-S / S-to-P $(\mathrm{cm}) \downarrow$ & NC $\uparrow$ & f-score $\uparrow$  
                                                                   & CD: P-to-S / S-to-P $(\mathrm{cm}) \downarrow$ & NC $\uparrow$ & f-score $\uparrow$  \\

    \midrule



    Our two-process framework       & $1.661/1.783$ & 0.846 & 41.651 & $1.466/1.677$ & 0.830 & 45.635 \\ 

    +Supervisor regularization  & $1.624/1.739$ & 0.850 & 42.204 & $1.414/1.641$ & 0.835 & 46.086 \\    
    +Cascading training        & $1.609/1.699$ & 0.851 & 42.822 & $1.396/1.616$ & 0.836 & 46.524 \\
    +LBS augmentation  & $1.602/1.726$ & 0.847 & 42.390 & $1.449/1.669$ & 0.832 & 45.342 \\    

    +Animation augmentation  & $1.546/1.657$ & 0.857 & 44.426 & $1.339/1.570$ & 0.840 & 47.605 \\    

    \bottomrule                     
    \end{tabular*}
    }

\end{table*}

\section{Experiments}
\label{sec:exp}

\subsection{Datasets and Evaluation Metric}
\textbf{Datasets.} 
To fairly compare our approach with existing methods, we followed the settings of the previous works, ICON~\cite{xiu2022icon} and SiTH~\cite{ho2024sith}, to train our model on the publicly available 3D human dataset, Thuman2.0~\cite{tao2021function4d_thuman}. Moreover, we followed the method of MultiGO~\cite{zhang2024multigo} to evaluate the effectiveness of our method on two test sets: CustomHumans~\cite{ho2023customhuman} and THuman3.0~\cite{thuman3.0}, allowing for a comparison with SOTA methods. To train our animation model, we prepared our triplet data from the Xhuman dataset~\cite{shen2023xavatar}. In the evaluation stage, all methods utilized the estimated SMPL/SMPL-X models as the body prior instead of the GT ones. 


\noindent \textbf{Evaluation Metrics.} Building upon previous studies, like SiTH~\cite{ho2024sith} and MultiGO~\cite{zhang2024multigo}, we utilize widely accepted 3D and 2D metrics in the field to conduct a quantitative comparison of our method with other SOTA approaches. For the 3D metrics, we calculate Chamfer distance (CD), Normal Consistency (NC), and f-score~\cite{fscore} to assess the performance of the methods. In terms of 2D metrics, we evaluate the methods using LPIPS~\cite{lpips}, SSIM, and PSNR.



\begin{table}[!t] 
\caption{\textbf{More Ablation on CustomHuman. For united geometry, we ablate the impact of different geometry prior models on results. For supervisor regularization, We compare the impact of regularizing different intermediate hidden layer features on results. For animation augmentation, we ablate the impact of offline and online augmentation on results.}  \label{abl_exp_3d_more}}
    		\vspace{-0.3cm}

\begin{center}
\scalebox{0.72}{
\begin{tabular}
{l|ccc}
\toprule

Methods                                                        & CD: P-to-S / S-to-P $(\mathrm{cm}) \downarrow$  & NC $\uparrow$ & f-score $\uparrow$ \\


\midrule

 \multicolumn{4}{c}{United geometry} \\

\midrule
    Our two-process framework  & $1.661/1.783$ & 0.846 & 41.651  \\
    Ours (w/o SMPL) & $1.709/1.839$ & 0.840 & 41.017 \\
    Ours (w/o SMPL\&back normal) & $1.849/2.153$ & 0.831 & 37.586 \\

\midrule

 \multicolumn{4}{c}{Supervisor regularization (reg.)} \\

\midrule

    Our two-process framework & $1.661/1.783$ & 0.846 & 41.651 \\ 
    Ours (w/ All-block reg.)  & $1.642/1.756$ & 0.849 & 41.985 \\ 
    Ours (w/ Mid\&Up-block reg.)  & $1.624/1.739$ & 0.850 & 42.204 \\ 
\midrule

 \multicolumn{4}{c}{Animation augmentation (aug.)} \\
\midrule

     Ours & $1.609/1.699$ & 0.851 & 42.822  \\
     Ours (Offline, 10-time aug.) & $1.607/1.695$ & 0.851 & 42.845  \\
     Ours (Online) & $1.546/1.657$ & 0.857 & 44.426  \\



\bottomrule
    \end{tabular}
}

\end{center}
    		\vspace{-0.1cm}

\end{table}


\begin{table}[!t]
\caption{\textbf{Ablation of the Supervisor Regularization and Animation Augmentation in Texture Quality.} \label{abl_exp_2d}}
    		\vspace{-0.3cm}

\begin{center}
\scalebox{0.67}{
\begin{tabular}
{l|ccc}
\toprule

Methods  &LPIPS: F/B $\downarrow$ & SSIM: F/B $\uparrow$ & PSNR: F/B $\uparrow$  \\
\midrule

& \multicolumn{3}{c}{CustomHuman}  \\
\midrule
    Ours  & $0.0358/0.0577$ & $0.9701/0.9499$ & $23.3510/21.5149$ \\
    Ours (w/o Animation) & $0.0373/0.0585$ & $0.9674/0.9476$ & $23.0797/21.2725$ \\
    Ours (w/o Animation\&Supervisor)& $0.0386/0.0604$ & $0.9666/0.9446$ & $22.8969/21.0956$ \\

\midrule

& \multicolumn{3}{c}{THuman3.0} \\

\midrule
    Ours & $0.0398/0.0588$ & $0.9738/0.9552$ & $25.1805/23.1188$ \\ 
    Ours (w/o Animation) & $0.0415/0.0605$ & $0.9714/0.9539$ & $24.8252/23.0113$ \\
    Ours (w/o Animation\&Supervisor)& $0.0425/0.0613$ & $0.9696/0.9524$ & $24.5156/22.8495$ \\

\bottomrule
    \end{tabular}
}

\end{center}
    		\vspace{-0.1cm}

\end{table}

\subsection{Quantitative Comparison with SOTA}


Our method is compared with SOTA methods in Tables ~\ref{main_exp_3d} and ~\ref{main_exp_2d}. We take results mainly from MultiGO~\cite{zhang2024multigo}. Moreover, we reproduce PSHuman~\cite{li2024pshuman} and Human-3Diffusion~\cite{3diffusion} for the results.


\noindent \textbf{3D Comparison.} Table~\ref{main_exp_3d} shows that our method exhibits strong competitiveness compared to existing methods. Under the same experimental setup, our method showed consistent improvements in all 3D metrics, especially in the f-score metric, where our method outperforms the strongest SOTA, MultiGO, by 2.011 and 1.514 on the Custom Human and THuman3.0 datasets, respectively.

\noindent \textbf{2D Comparison.} Table~\ref{main_exp_2d} shows the advantages of our method over existing methods on textures. Our method showed improvements on both test sets. Specifically, compared to the strongest SOTA, multiGO, our method improves the SSIM metric by $0.98\%/0.84\%$(F/B) and $1.15\%/0.4\%$(F/B) on the CustomHuman and THuman3.0.

\begin{figure*}[t!]
    \centering
    \setlength{\belowcaptionskip}{-0.25cm}
    \includegraphics[width=1\linewidth]{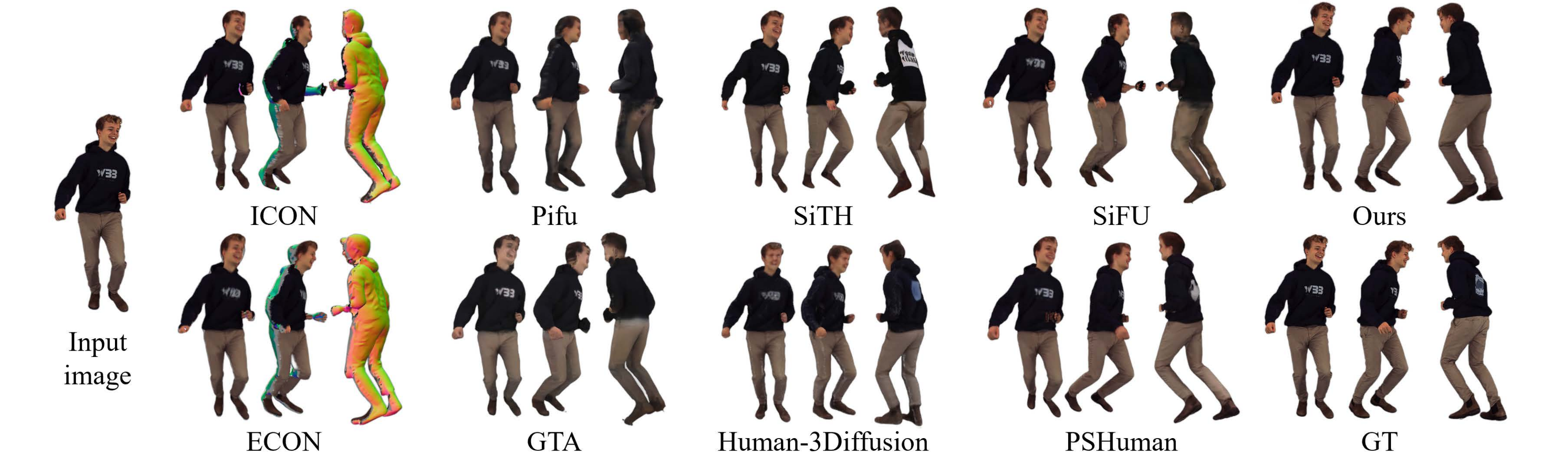}
    		\vspace{-0.8cm}
    \caption{\textbf{Qualitative Texture Comparison with SOTA Methods.} For the 3D result of each method, we render them from three views. Comparison finds our method can better reconstruct the human texture, and produces less blurring and limb deformities.}
    \label{fig:texture}
    \vspace{-0.2cm}

\end{figure*}

\begin{figure*}[t!]
    \centering
    \setlength{\belowcaptionskip}{-0.25cm}
    \includegraphics[width=1\linewidth]{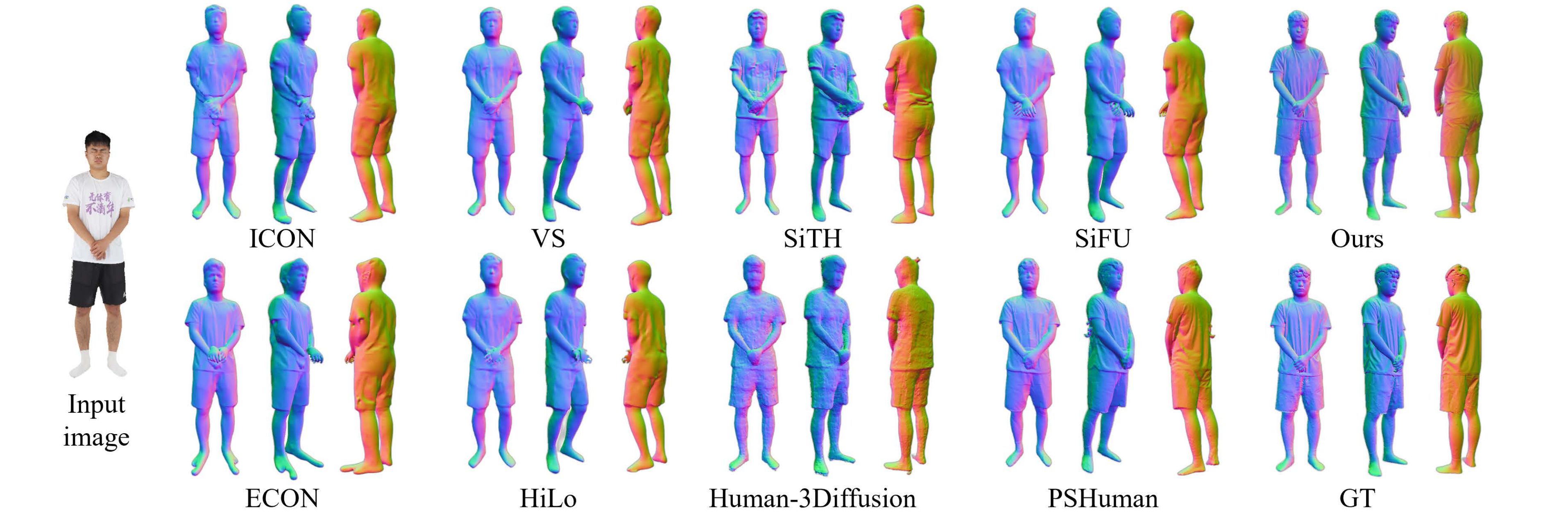}
    		\vspace{-0.8cm}
    \caption{\textbf{Qualitative Geometry Comparison with SOTA Methods.} From the geometry comparison with other methods, it is found that our method can also reconstruct more accurate and detailed 3D human geometry with less distortion.}
    \label{fig:geometry}

\end{figure*}

\subsection{Quantitative Ablation Study}
\label{sec:ablation}

\noindent \textbf{Component Ablation.} We provide a quantitative evaluation of each component we propose in Table~\ref{abl_exp_3d}. From Table~\ref{abl_exp_3d}, it is found that as we continuously increase the components we propose from top to bottom, the performance of our method improves accordingly. 

Adding our supervisor regularization into the baseline results in improvements. This arises from the fact that the trained supervisor model uses GT's human normal maps. So, it not only reconstruct the almost perfect 3D human,  but ensures that the feature outputted from hidden blocks in the network encapsulate the necessary target features for reconstruction. We use these features to constrain the corresponding-block features of the monocular reconstruction network. This approach can better help the monocular reconstruction predict the required features and improve the final results.

Moreover, the introduction of the proposed cascading training operation can also achieve certain performance improvements. Compared to training two processes separately (noted in Section~\ref{sec:CGT}), this method takes into account the setting during actual inference, allowing the second process to fully fit the output distribution of the first stage during training, thereby improving performance.

\begin{figure*}[t]
    \centering
    \setlength{\belowcaptionskip}{-0.25cm}
    \includegraphics[width=0.95\linewidth]{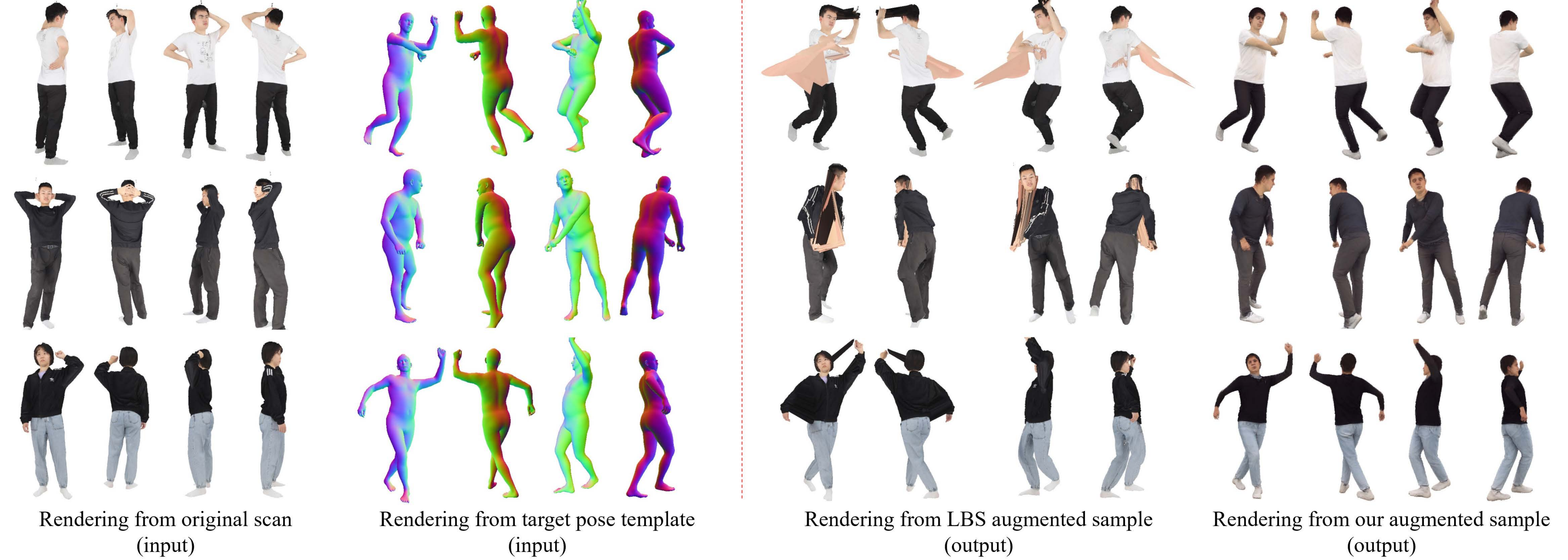}
    \caption{\textbf{Comparison of Different Online Augmented Samples.} Compared to LBS augmented samples, our augmented samples do not produce any distortion, which can promote the model rather than damage its original performance. It is worth mentioning that our animation model generates 3D human data with some differences from original scans, but these differences don't affect their suitability as reconstruction training samples. For reconstruction, they are merely equivalent to other people. Maintaining human ID consistency after animation is crucial for animation field but beyond the reconstruction research scope.}
    \label{fig:Ani_abl}

\end{figure*}

From the comparison of the last two rows in Table~\ref{abl_exp_3d}, it is found that our proposed augmentation method is more effective compared to the LBS method. Although both are online augmentation methods, introducing samples generated from the LBS method can lead to a decrease in performance (The fourth row compares to the third). But our method can achieve an improvement. This is because, compared to the samples from our method, the quality of the samples generated by LBS is worse. Readers might refer to Figure~\ref{fig:Ani_abl} for intuitive comparison.

\begin{figure}
    \centering

    \setlength{\belowcaptionskip}{-0.1cm}
    \includegraphics[width=1\linewidth]{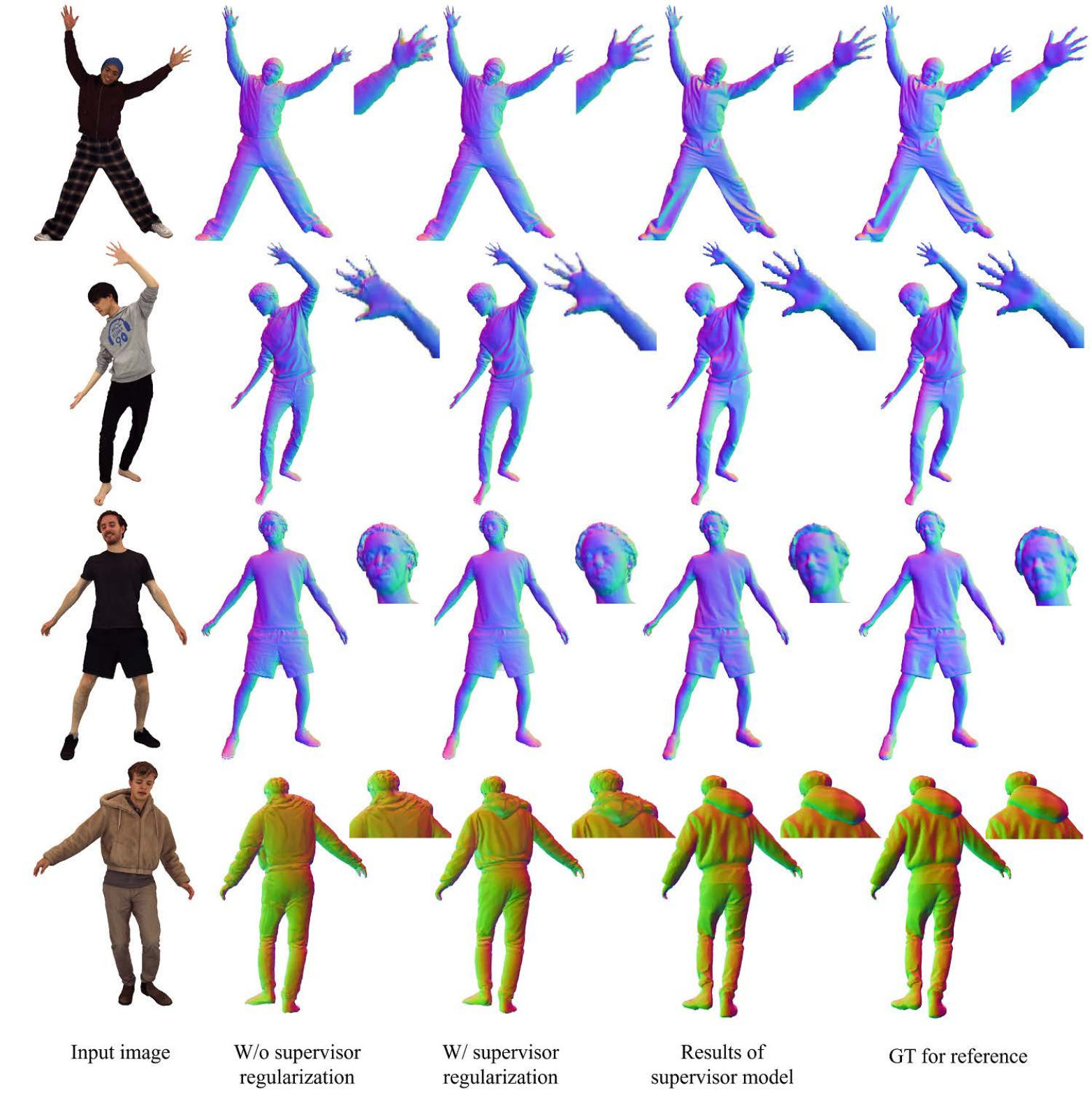}
    		\vspace{-0.75cm}

    \caption{\textbf{Visual Impact of Supervisor Regularization.} It is found that the supervisor model can reconstruct 3D human almost the same as GT. After using our supervisor regularization, the reconstructed details of human can be improved. }
    \label{fig:Sup_abl}

\end{figure}

\noindent \textbf{More Ablation.} In Table~\ref{abl_exp_3d_more}, we ablate the components we propose in detail. From ablation of united geometry, it is found that the input of different geometric priors will gradually improve the final reconstruction results. For supervisor regularization, it is experimentally found that it is more effective when only the $mid$ and $up$ blocks of features are supervised. The possible reason is that due to the difference in input between the supervisor and the monocular reconstruction network, there is a great difference in the shallow reconstruction features. Forcing constraints may interfere with the feature extraction of the monocular reconstruction model. For animation augmentation, it is found that augmenting samples offline (We enhance each original scan 10 times and save them locally) does not notably improve the performance. This is because, compared to the data size augmented by online learning, the samples generated offline are limited. Meanwhile, another advantage of online learning is that it requires fewer local resources and is more efficient. In Table~\ref{abl_exp_2d}, we also ablated the proposed supervisor module and animation module on the 2D metric, and obtained the same conclusion. As we gradually removed these two components, there was a gradual decline in performance.

\subsection{Visualiztion}
\label{sec:vis}

\textbf{Comparison with SOTA Methods.} Figures~\ref{fig:teaser},~\ref{fig:texture} and~\ref{fig:geometry} show the advantages of our method compared to the current SOTA methods. In contrast, the 3D human reconstructed by our method is closer to GT, with fewer texture blurring such as facial and clothing printing, and geometric distortions such as tattered hands.

\noindent \textbf{Comparison of Augmented Samples.} Figure~\ref{fig:Ani_abl} shows the advantages of our augmented samples compared to those from LBS method. It is found that the samples from the LBS method exhibit significant distortion. This explains why introducing these samples leads to a performance decrease in Table~\ref{abl_exp_3d}. In contrast, our animation model is capable of performing excellent transformations to specified poses and generating samples without any distortion.

\noindent \textbf{Comparison of Using/Not Using Supervisor Regularization.} Figure~\ref{fig:Sup_abl} shows the difference before and after introducing our supervisor regularization. It is found that with supervisor regularization, the model can achieve better reconstruction results for details of the human body, such as fingers, facial details, and clothing hoods.

\section{Conclusion}
\label{sec:conclusion}
In conclusion, the proposed SAT framework effectively advances monocular texture 3D human reconstruction through its innovative two-process approach. The integration of UGL and CGT processes enables precise human geometry and texture reconstruction by utilizing diverse geometric priors and avoiding cascading errors. The introduction of the SFR module ensures enhanced accuracy in geometric learning, while the OAA module addresses data scarcity by generating augmented samples online. Our method demonstrates SOTA performance on public datasets, validating its contribution.

    



\bibliographystyle{ACM-Reference-Format}
\bibliography{sample-base}

\end{document}